\documentclass[10pt,journal,twoside,print]{ieeecolor}

\usepackage{generic}
\usepackage{cite}
\usepackage{amsmath,amssymb,amsfonts}
\usepackage{algorithmic}
\usepackage{graphicx}
\usepackage{textcomp}
\usepackage{bm}
\usepackage{subcaption}
\let\proof\relax
\let\endproof\relax

\usepackage{amsthm}
\newtheorem{theorem}{Theorem}
\usepackage{siunitx}
\newtheorem{definition}{Definition}[section]
\newtheorem{remark}{Remark}
\newtheorem{assumption}{Assumption}

\DeclareMathOperator*{\diag}{diag}

\usepackage{lipsum}
\usepackage{titlesec}

\titlespacing\section{0pt}{2pt plus 2pt minus 0pt}{0pt plus 2pt minus 0pt}
\titlespacing\subsection{0pt}{2pt plus 2pt minus 0pt}{0pt plus 2pt minus 0pt}

\def\BibTeX{{\rm B\kern-.05em{\sc i\kern-.025em b}\kern-.08em
    T\kern-.1667em\lower.7ex\hbox{E}\kern-.125emX}}
\makeatletter
\let\NAT@parse\undefined
\makeatother

\usepackage[pdftex, plainpages=false,hypertexnames=true,pdfnewwindow=true,colorlinks=true,citecolor=blue,linkcolor=red,urlcolor=blue,filecolor=blue]{hyperref}%

\begin{document}

\title{\LARGE \bf Input State Stability of Gated Graph Neural Networks}

\author{Antonio Marino$^{1}$, Claudio Pacchierotti$^{2}$, Paolo Robuffo Giordano$^{2}$ 
\thanks{$^{1}$ A. Marino is with Univ Rennes, CNRS, Inria, IRISA -- Rennes, France. E-mail: antonio.marino@irisa.fr}
\thanks{$^{2}$ C. Pacchierotti and P. Robuffo Giordano are with CNRS, Univ Rennes, Inria, IRISA -- Rennes, France. E-mail: \{claudio.pacchierotti,prg\}@irisa.fr.}
\thanks{This work was supported by the ANR-20-CHIA-0017 project ``MULTISHARED''}
}

\maketitle

\begin{abstract}
In this paper, we aim to find the conditions for input-state stability (ISS) and incremental input-state stability ($\delta$ISS) of Gated Graph Neural Networks (GGNNs). We show that this recurrent version of Graph Neural Networks (GNNs) can be expressed as a dynamical distributed system and, as a consequence, can be analysed using model-based techniques to assess its stability and robustness properties. Then, the stability criteria found can be exploited as constraints during the training process to enforce the internal stability of the neural network. Two distributed control examples, flocking and multi-robot motion control, show that using these conditions increases the performance and robustness of the gated GNNs.
\end{abstract}

\begin{IEEEkeywords}
 distributed control, graph neural network, stability analysis
\end{IEEEkeywords}

\section{INTRODUCTION}
 \IEEEPARstart{I}{n} recent years, multi-agent systems have garnered considerable attention~\cite{olfati2007consensus}. Coordinated multi-agent systems offer enhanced collaborative problem-solving capabilities and flexibility compared to single-agent approaches, making them suitable for various disciplines such as computer science, electrical engineering, and robotics~\cite{dorri2018multi}. Effective coordination of multiple agents requires considering group dynamics. In the case of multi-robot applications, generating individual robot motion involves not only local sensing data, but also information about the group state, often acquired through communication with a limited number of neighboring team members~\cite{I-Magnus2017ControlMRSsurvey}. Thus, communication becomes a critical element in realizing distributed solutions for multi-agent systems.

Neural networks have gained prominence in data-driven control applications, given their exceptional approximation capabilities~\cite{huang2006universal}. In distributed control, neural networks are particularly advantageous as they can approximate complex distributed policies without intricate optimizations and designs. The literature showcases various data-driven approaches, especially those rooted in reinforcement learning~\cite{batra2022decentralized, he2020integral}, leveraging input data like images~\cite{gupta2017cooperative}. However, these approaches operate on local sensing data without inter-agent communication. On the other hand, distributed machine learning employs communication to partition the learning process across multiple machines contributing to group knowledge~\cite{majcherczyk2021flow}.

Leveraging communication, recent trends in data-driven distributed control involve employing Graph Neural Networks (GNNs) to encode distributed control and planning solutions. GNNs excel in predicting and analyzing graphs, proving valuable in various problems such as text classification, protein interface predictions, and social network decisions~\cite{I-Lingfei2022GNNFoundations}. Particularly in the latter cases, GNNs outperform classical neural network architectures~\cite{I-Jegelka2019GNNpowerful}, offering a fresh perspective on distributed control implementation.
In this respect, Gama et al.~\cite{gama2021graph} extend GNNs to flocking control for large teams. Additional examples of the use of GNNs for distributed control are found in space coverage~\cite{li2021message}, multi-robot path planning~\cite{tolstaya2021multi}, and motion planning~\cite{khan2020graph}, including obstacle-rich environments~\cite{ji2021decentralized}. GNNs also enhance multi-agent perception~\cite{zhou2022multi} and enable distributed active information acquisition~\cite{P-Zhou2022GNNPerception}, translating multi-robot information gathering into graph representation and formulating GNN-based decision-making.

In the recent literature, a very relevant discussion is about making learning based method robust and stable~\cite{hewing2020learning}. In this context, many works applied contraction analysis to demonstrate recurrent neural network stability~\cite{davydov2022non}, or directly closed-loop stability in continuous learning~\cite{song2022stability} and adaptive control~\cite{tsukamoto2021learning}.
Recently, the stability analyses presented in~\cite{bonassi2021stability,terzi2021learning} showcased the concepts of ISS (Input-to-State Stability) and incremental ISS ($\delta$ISS)~\cite{bayer2013discrete} in the context of LSTMs and GRUs, which are two of the most popular recurrent neural network models. The $\delta$ISS property is a stronger property than plain ISS as it leads to the asymptotically convergence of two state trajectories when their respective input sequences are close, regardless of the initial conditions for the states. In a broader scope, D'Amico et al.~\cite{d2022incremental} has established the criteria for achieving $\delta$ISS in a general class of recurrent neural networks. Additionally, they have formally outlined the Linear Matrix Inequality (LMI) problem that guarantees the stability of the acquired neural network.

Inspired by these last results, this work characterizes the $\delta$ISS properties of the recurrent version of GNN, i.e. Gated Graph Neural Networks (GGNN)~\cite{ruiz2020gated}. These models use gated mechanisms to deploy distributed recurrent models able to reason on temporal- and spatial-based relationships among the agents. Compared to the recurrent systems in~\cite{d2022incremental}, GGNNs present a heightened level of complexity. This is primarily due to the fact that their underlying communication graph is not predetermined and undergoes changes over time, ultimately rendering the entire neural network variant across time.

To the best of our knowledge, this is the first time ISS and $\delta$ISS have been proven for GGNN. Previous works have primarily focused on limited stability results, such as stability to graph perturbations~\cite{gama2020stability, ruiz2020gated}. Instead, this article considers the system internal stability to the input features in a more general dynamical system analysis. The conditions derived in this work are in the form of nonlinear inequalities on GGNN weights: these can be exploited to certify the stability of a trained neural model, or can be enforced as constraints during the training process to guarantee the stability of the GGNN.

The remaining sections of this paper are organized as follow. Section~\ref{Preliminaries} reviews the preliminaries about graph neural networks and presents their recurrent versions, RGNNs and GGNNs. Section~\ref{one-ggnn-layer} and~\ref{ggnn-stability-deep} present the main stability results for the one-layer and deep GGNN. In Sect.~\ref{ggnn-stability-delay}, we also provide a perspective of these results when considering communication delays. We then show two examples of the conditions found for flocking control, in Sect.~\ref{flocking-control-example}, and multi robot motion control, in Sect.~\ref{multi-robot-motion-planning-example}. Concluding remarks are finally presented in Sect.~\ref{conclusions}.

\section{PRELIMINARIES}
\label{Preliminaries}
Let $\mathcal{G}=(\mathcal{V},\,\mathcal{E})$ be an undirected graph where $\mathcal{V} = \{v_1, \dots, v_N\}$ is the vertex set (representing the $N$ agents in the group) and $\mathcal{E}\subseteq \mathcal{V}\times \mathcal{V}$ is the edge set. 
Each edge $e_k=(i,\,j)\in\mathcal{E}$ is associated with a weight $w_{ij}\geq 0$ such that $w_{ij}>0$ if the agent $i$ and $j$ can interact and $w_{ij}=0$ otherwise. 
As usual, we denote with $\mathcal{N}_i=\{j\in\mathcal{V}|\;w_{ij}>0\}$ the set of neighbors of agent $i$.
We also let $\bm{A} \in \mathbb{R}^{N\times N}$ be the adjacency matrix with entries given by the weights $w_{ij}$. Defining the degree matrix $\bm{D} =\diag(d_i)$ with $d_i = \sum_{j\in \mathcal{N}_i} w_{ij}$, the Laplacian matrix of the graph is $\bm{L}=\bm{D}-\bm{A}$.

The graph signal $\bm{x} \in \mathbb{R}^N$, whose $i$-component $x_i$ is assigned to agent $i$, can be processed over the network by the following linear combination rule applied by each agent
\begin{equation}
    \small
    \boldsymbol s_i\bm{x} = \sum_{j\in \mathcal{N}_i} s_{ji} (x_i - x_j),
    \label{eq:aggregation}
\end{equation}
where $\boldsymbol s_i$ is the $i$-th row of $\bm{S}$. The signal manipulation can be operated by means of any \textit{graph shift operator} $\bm{S}$, e.g., Laplacian, adjacency matrix, weighted Laplacian , which respects the sparsity pattern of the graph. Later, in Sects.~\ref{flocking-control-example},~\ref{multi-robot-motion-planning-example}, we will use the Laplacian as support matrix, as it is commonly used in distributed control. However, the proposed techniques do not assume the use of a specific support matrix.

Performing $k$ repeated applications of $\bm{S}$ on the same signal represents the aggregation of the $k$-hop neighborhood information. In analogy with traditional signal processing, this property can be used to define a linear graph filtering~\cite{P-Shuman2013SPG} that processes the multi features signal $\bm{x} \in \mathbb{R}^{N \times G}$ with $G$ features:
\begin{equation}
\small
    H_{\bm{S}}(\bm{x}) = \sum_{k=0}^K \bm{S}^k \bm{x} \bm{H_k}.
    \label{eq:filter}
\end{equation}
where the weights $\bm{H_k} \in \mathbb{R}^{G \times F}$ define the output of the filter. Note that $\bm{S}^k=\bm{S}(\bm{S}^{k-1})$, so that it can be computed locally with repeated 1-hop communications between a node and its neighbors. Hence, the computation of $H_{\bm{S}}$ is distributed on each node.
\subsection{Graph Neural Network}
\label{P-GNN}
Although $H_{\bm{S}}$ is simple to evaluate, it can only represent a linear mapping between input and output filters. GNNs increase the expressiveness of the linear graph filters by means of pointwise nonlinearities $\rho : \mathbb{R}^{N \times F_{l-1}} \rightarrow{} \mathbb{R}^{N \times F_{l-1}}$ following a filter bank. Letting $H_{\bm{S}l}$ be a bank of $F_{l-1} \times F_l$ filters at layer $l$, the GNN layer is defined as
\begin{equation}
    \bm{x}_l = \rho(H_{\bm{S}l}(\bm{x}_{l-1})), \qquad  \bm{x}_{l-1} \in \mathbb{R}^{N \times F_{l-1}}.
\end{equation}
Starting by $l=0$ with $F_0$, the signal tensor $\bm{x}_{l_n} \in \mathbb{R}^{N \times F_{l_n}}$ is the output of a cascade of $l_n$ GNN layers. This specific type of GNN is commonly referred to as a convolution graph network because each layer utilizes a graph signal convolution~\eqref{eq:filter}. By the use of imitation learning or similar techniques, the GNN can learn a distributed policy by finding the optimal filter weights $\bm{H_k}$ to propagate information among the agents and generate the desired output. Notably, each agent employs an identical version of the network and exchanges intermediate quantities with the other agents in the network through the application of $S$, resulting in an overall distributed neural network.
GNNs inherit some interesting properties from graph filters, such as permutational equivariance~\cite{gama2020stability} and their local and distributed nature, showing superior ability to process graph signals~\cite{P-Zhou2022GNNPerception,P-Qingbiao2020GNNPathPlanning,P-Tolstaya2021Coverage}. Under dynamic graphs and perturbations of $\bm{S}$, the stability of GNN predictions is related to $\bm{S}$ spectral characteristics. In particular, the graph filters composing the GNN have a frequency response,  $h(\lambda)$~\cite{P-Sandryhaila2014FrequenciesSPG}, bounded by the graph support eigenvalues $\lambda$ if the graph filters are integral Lipschitz:
\begin{equation}
\small
    |h(\lambda_j)-h(\lambda_i)| \leq 2C \frac{|\lambda_j-\lambda_i|}{|\lambda_j+\lambda_i|}
    \label{eq:GNN-stability}
\end{equation}
where $\lambda_j,\lambda_i \in \mathbb{R}$ are any support matrix eigenvalues, and $C>0$ a proper integral Lipschitz constant. This condition restricts the graph frequency response variability to the midpoint of $\lambda$ variations. 

\subsection{Gated Graph Neural Network}
\label{P-GGNN}
Recurrent models of GNNs can solve time-dependent problems. These models, similarly to recurrent neural networks (RNNs), are known as graph recurrent neural networks (GRNNs). GRNNs utilize memory to learn patterns in data sequences, where the data is spatially encoded within graphs, regardless of the team size of the agents~\cite{gu2020implicit}. However, traditional GRNNs encounter challenges such as vanishing gradients, which are also found in RNNs. Additionally, they face difficulties in handling long sequences in space, where certain nodes or paths within the graph might be assigned more importance than others in long-range exchanges, causing imbalances in the graph's informational encoding. \\
Forgetting factors can be applied to mitigate this problem, reducing the influence of past or new signal on the state. A Gated Graph Neural Network (GGNN)~\cite{ruiz2020gated} is a recurrent Graph Neural Network that uses a gating mechanism to control how the past information influences the update of the GNN states. We can add a state and an input gate, $\bm{\hat{q}}, \bm{\tilde{q}} \in Q \subseteq [0,1]^{N \times F}$, that are multiplied via the Hadamard product~$\circ$ by the state and the inputs of the network, respectively. These two gates regulate how much the past information or the input are used to update the network internal state. GGNNs admit the following state-space representation~\cite{lukovnikov2020improving}, 
\begin{equation}
\small
    \begin{cases}
    \tilde{\bm{q}} = \sigma(\tilde{A}_S(\bm{x}) + \tilde{B}_S(\bm{u}) +\hat{b}) \\
    \hat{\bm{q}} = \sigma(\hat{A}_S(\bm{x}) + \hat{B}_S(\bm{u}) +\tilde{b})  \\
    \bm{x}^+ = \sigma_c(\bm{\hat{q}}\circ A_S(\bm{x}) + \bm{\tilde{q}} \circ B_S(\bm{u}) +b)
    \end{cases}
    \label{system}
\end{equation}
with $\sigma(x) = \frac{1}{1+e^{-x}}$ being the logistic function, and $\sigma_c(x) = \frac{e^{x}-e^{-x}}{e^{x}+e^{-x}}$ being the hyperbolic tangent. $\hat{A}_S,\hat{B}_S$ are graph filters of the forgetting gate, $\tilde{A}_S,\tilde{B}_S$ are graph filters~\eqref{eq:filter} of the input gate, and $A_S$ and $B_S$ are the state graph filters~\eqref{eq:filter}. $\bm{\hat{b}},\bm{\tilde{b}},\bm{b} \in \mathbb{R}^{N \times F}$ are respectively the biases of the gates and the state built as $\bm{1}_N \otimes \bm{\textit{b}}$ with the same bias for every agents. The weights and biases of the graph filters can be acquired using imitation learning techniques~\cite{ross2011reduction}

We note that, in the literature, there exist several Gated GNNs structures \cite{lukovnikov2020improving,ruiz2020gated,I-Lingfei2022GNNFoundations}, sharing gating mechanisms of different kinds (temporal, attention, and so on). It is easy to check that the system in~\eqref{system} generalizes the systems above, thereby offering a broader framework for analyzing their dynamic stability. We want to characterize the dynamic stability properties of such neural models that combine gating mechanisms with a distributed computation. Therefore, we have to analyze the state $\bm{x}$ evolution and its convergence as a nonlinear dynamical system.

\subsection{Incremental Input State Stability}
\label{P-dISS}
In this section, we recall the definition of ISS and $\delta$ISS that will be used throughout the paper. The ISS property guarantees that, regardless of the starting conditions, limited inputs or disturbances lead to limited system states. This attribute is advantageous in control systems, as they require stability and robustness. However, this principle does not hold for systems that are inherently unstable. Recalling the definitions of $\mathcal{KL}$, $\mathcal{K}_\infty$ functions~\cite{jiang2001input}, and the infinite norm $|| \cdot||_{\infty}$, the following definition of ISS is given
\begin{definition}[ISS]
\label{ISS_def}
System~\eqref{system} is called input-to-state stable if there exist functions $\beta \in \mathcal{KL} $ and $\gamma \in \mathcal{K}_\infty$ such that, for any $t \in \mathbb{Z} \geq 0$, any initial state $\bm{x}(0) \in \mathcal{X}$ any input sequence $\bm{u} \in \mathcal{U}$ it holds that:
\begin{equation}
\small
    ||\bm{x}(t)||_{\infty} \leq \beta(||\bm{x}(0)||_{\infty},t) + \gamma_u(||\bm{u}||_{\infty}) + \gamma_b(||\bm{b}||_{\infty})
\end{equation}

\end{definition}

\noindent A further desirable property is incremental ISS ($\delta$ISS)~\cite{angeli2002lyapunov}. The $\delta$ISS property ensures that any pair of state trajectories converge towards each other even if they start from different initial conditions. Moreover, their difference is bounded only by the differences of their inputs (e.g., an ideal control corrupted by an additive noise), thus enhancing the system robustness~\cite{jouffroy2003simple}.     

\begin{definition}[$\delta$ISS]
\label{dISS_def}
System~\eqref{system} is called incrementally input-to-state stable~\cite{bayer2013discrete} if there exist functions $\beta_{\delta} \in \mathcal{KL} $ and $\gamma_{\delta} \in \mathcal{K}_\infty$ such that, for any $t \in \mathbb{Z} \geq 0$, any initial states $\bm{x}(0)_1,\bm{x}(0)_2  \in \mathcal{X}$ any input sequences $\bm{u_1},\bm{u_2} \in \mathcal{U}$ it holds that:
\begin{equation}
\small
    \begin{aligned}
    ||\bm{x}(t)_1 - \bm{x}(t)_2||_{\infty} \leq & \beta_{\delta}(||\bm{x}(0)_1 - \bm{x}(0)_2 ||_{\infty},t)  \\ &+ \gamma_{\delta}(||\bm{u}_1 - \bm{u}_2||_{\infty})
    \end{aligned}
\end{equation}
\end{definition}
\begin{remark} 
In the neural network context, the $\delta$ISS property ensures that any difference in the initial conditions will be eventually discarded, and thus the same outputs will correspond to the same observations. Moreover, since the stability is valid for $t>0$, for a training with a finite time sequence dataset it is guaranteed that all the NN state trajectories converge to a unique solution.
\end{remark}

\section{ONE-LAYER GGNN STABILITY}
\label{one-ggnn-layer}
In this section we will discuss the stability properties of a single layer GGNN. For the rest of the paper the following assumption will be made
\begin{assumption}
The input $\bm{u}$ is unity-bounded: $\bm{u} \in \mathcal{U} \subseteq [-1,1]^{N \times G}$ , i.e. $||\bm{u}||_{\infty} \leq 1$.
\label{assumption1}
\end{assumption}
\noindent This is a quite mild assumption since the input signal is usually normalized or it is the result of others network layers with unitary output activation functions. \\
Before stating the sufficient conditions for the ISS of GGNN, we will first introduce the notation for the following quantities
\begin{equation}
\small
    \begin{aligned}
        & \qquad S_K \triangleq [I, S,\dots, S^K ] \\ 
        A   \triangleq [A_0, \dots, A_K ]^T & \qquad
        B  \triangleq [B_0, \dots, B_K ]^T \\
        \tilde{A} \triangleq [\tilde{A}_0, \dots, \tilde{A}_K]^T & \qquad
        \hat{A}  \triangleq [\hat{A}_0, \dots, \hat{A}_K]^T  \\
        \tilde{B}  \triangleq [\tilde{B}_0, \dots, \tilde{B}_K ]^T & \qquad
        \hat{B} \triangleq [\hat{B}_0, \dots, \hat{B}_K ]^T
    \end{aligned}
    \label{eq:definitions}
\end{equation}
where $K$ is the filters length. Then, in light of assumption~\eqref{assumption1} and knowing that $ ||\bm{x}||\leq 1$,  each gate feature $q_i$ satisfies:

\begin{equation}
\small
    \begin{aligned}
    |\hat{q}_i| & \leq || \bm{\hat{q}} ||_{\infty} \leq \max_{u \in \mathcal{U}, x \in \mathcal{X} } || \sigma(\hat{A}_S(\bm{x}) + \hat{B}_S(\bm{u}) + \hat{b})||_\infty \\ & \leq || \max_{u \in U, x \in \mathcal{X} } \sigma(\hat{A}_S(\bm{x}) + \hat{B}_S(\bm{u}) + \hat{b})||_\infty \\ & \leq \sigma(  \max_{u \in U, x \in \mathcal{X} } ||\hat{A}_S(\bm{x}) + \hat{B}_S(\bm{u}) + \hat{b}||_\infty) \\ & \leq \sigma(|| \hat{A}_S(\bm{x}_{max}) + \hat{B}_S(\bm{u}_{max})+ \hat{b} ||_\infty)
    \end{aligned}
    \label{eq:gates_maj}
\end{equation}
where $\bm{x}_{max}$ and $\bm{u}_{max}$ are the maximum values of $\bm{x}$ and $\bm{u}$. Recalling that a graph filter in equation~\eqref{eq:filter} can be written as $\hat{H}_S(\bm{x})=[I ,S,\dots,S^K]( I_{K} \otimes \bm{x})[\hat{H}_0,\hat{H}_1,\dots,\hat{H}_K]^T$ and using notation~\eqref{eq:definitions}, equation~\eqref{eq:gates_maj} becomes 
\begin{equation}
\small
    \begin{aligned}
        |\hat{q}_i| & \leq \sigma( ||S_K||_{\infty}(|| \hat{A} ||_{\infty} ||\bm{x}||_{\infty} + ||\hat{B}||_{\infty} ||\bm{u}||_{\infty})+||\hat{\bm{b}}||_\infty) \\ & = 
        \sigma( ||S_K||_{\infty}(|| \hat{A} ||_{\infty}+ ||\hat{B}||_{\infty})+||\hat{\bm{b}}||_\infty) \triangleq \sigma_{\hat{q}}.
    \end{aligned}
\label{eq:hat_gate}
\end{equation}
We identify the induced $\infty$-norm as $||\cdot||_{\infty}$. Similarly for $\tilde{q}_i$
\begin{equation}
\small
\begin{aligned}
    |\tilde{q}_i| & \leq \sigma( ||S_K||_{\infty}(||\tilde{A}||_{\infty}+|| \tilde{B}||_{\infty}) +||\tilde{\bm{b}}||_\infty) \triangleq \sigma_{\tilde{q}}.
\end{aligned}
\label{eq:tilde_gate}
\end{equation}

\begin{theorem}
\label{ISS_stab}
A sufficient condition for the ISS of a single-layer GGNN network is that $\mathcal{A} \leq 1$, where
\begin{equation}
\small
    \mathcal{A} \triangleq \sigma_{\hat{q}} ||S_K||_{\infty} ||A||_{\infty}.
\end{equation}
\end{theorem}
\noindent The proof based on the results for GRU and LSTM ~\cite{bonassi2021stability,terzi2021learning} is provided in the Appendix~\ref{proof_ISS}.

The $\delta$ISS stability requires to analyze the evolution of the maximum distance of two states trajectories $\bm{x_1}, \bm{x_2}$, starting from two different initial conditions $\bm{x}_1(0),\bm{x}_2(0)$ and having two different inputs $\bm{u_1},\bm{u_2}$. We also need another assumption
\begin{assumption}
\label{assumption2}
Given two support matrices $||S_{1}(t)||_{\infty},||S_{2}(t)||_{\infty},\forall t \in \mathbb{Z}^+$ associated with two different graphs, they are bounded by the same $||\bar{S}||_{\infty}$.
\end{assumption}
\begin{theorem}
\label{dISS_stab}
Under the assumption~\ref{assumption2}, a sufficient condition for the system~\eqref{system} to be $\delta$ISS is $\mathcal{A}_{\delta} \leq 1$; where
\begin{flalign}
    \begin{aligned}
    \mathcal{A}_{\delta} & \triangleq \sigma_{\hat{q}}||\bar{S}||_\infty ||A||_{\infty}+\frac{1}{4}||\bar{S}||^2_\infty||\hat{A}||_{\infty} ||A||_{\infty} \\ & + \frac{1}{4} ||\bar{S}||^2_\infty||\tilde{A}||_{\infty} ||B||_{\infty}.
    \end{aligned}
    \end{flalign}
\end{theorem}
\noindent The proof to this theorem is reported in  Appendix~\ref{prrof_dISS}. 
Assumption~\ref{assumption2} is reasonable for multi-agent systems, since graphs always have a finite number of agents with finite number of links between each other. 
When we deal with scalability and dynamic graphs, the assumption may be restrictive based on the choice of the support matrix. For examples, by using the adjacency matrix the upper bound of its norm would be the maximum number of the links for one agent, i.e. $N$ (the team size). Therefore, training the GGNN using the adjacency matrix and being $||\bar{S}||_{\infty} = N$, the stability condition would be respected for teams with a maximum number of links for each agent up to $N$. For group with $N' > N$ we can guarantee stability if the agents have a number of neighbors less than $N$. 

\begin{remark}
\label{remark-ver}
The condition presented in Theorem 2 also offers an approach for verifying the $\delta$ISS characteristic of a GGNN layer, in contrast of relying on a statistical analysis of state convergence. This condition, while sufficient, provides a computationally inexpensive tool for assessment.
\end{remark}
\begin{remark}
\label{remark-nL}
In practice we can solve this issue by constraining, at runtime, the cardinality of $\# \mathcal{N}_i < N$ for every agent $i$ in the team. For a GGNN that uses normalized support matrices, e.g., normalized Laplacian, the assumption is met without any further restrictions on the graph topology. Moreover the use of normalized support reduces the regularizing term in the loss function at training time. In the experiments we will use normalized Laplacian to compare stable and unstable neural network.
\end{remark}

\begin{remark}
Note that we did not make any assumptions on the shape of $S$; the graph can be made of connected or disconnected sub-graphs without affecting the stability.
\end{remark}
  While the identified conditions are deemed sufficient, it is important to acknowledge the possibility that certain GGNN layers may exhibit $\delta$ISS or ISS characteristics without necessarily conforming to the conditions outlined in Theorem~\ref{dISS_stab} or Theorem~\ref{ISS_stab}. In the upcoming sections, we will demonstrate that satisfying these conditions, although potentially restrictive, leads to enhanced performance.
\section{DEEP GGNN STABILITY}
\label{ggnn-stability-deep}
With the word ``deep" we refer to the stack of multiple layers of the Neural Networks. It is natural to ask whether the stability properties extend from the single layer to a multi-layer structure, commonly used in practice. A deep GGNN, whose representation is reported in the supplementary material, is made by interconnections among layers by feeding the future state of one layer to the next one,
\begin{subequations}
\begin{equation}
\small
    \begin{cases}
    \tilde{\bm{q}}^i = \sigma(\tilde{A}_S(\bm{x^i}+ \tilde{B}_S(\bm{u^i}) + \hat{\bm{b}}) \\
    \hat{\bm{q}}^i = \sigma(\hat{A}_S(\bm{x^i} + \hat{B}_S(\bm{u^i})+ \tilde{\bm{b}}))  \\
    \bm{x}^{i+} = \sigma_c(\hat{q}^i\circ A_S(\bm{x^i}) + \tilde{q}^i \circ B_S(\bm{u^i}) + \bm{b}) \\
    \bm{u^i} = \bm{x^{i-1 +}}, \quad \bm{u^1} = \bm{u}  
    \end{cases}
\end{equation}\\
for all the layers $i \in \{ 1, \dots, M\}$. The output of the network results from a graph output filter \\
\begin{equation}
\small
\bm{y} = Y_S(\bm{x^M}) + \bm{b_y}.
\end{equation}
\label{deep-GGNN}
\end{subequations}
The following then holds 
\begin{theorem}
The GGNN network is ISS if $\mathcal{A}^{i} \leq 1$ for every layer $i \in \{ 1, \dots, M\}$, where $\mathcal{A}^i$ are defined like in Theorem~\ref{ISS_stab}
\end{theorem}
\begin{proof}
The deep GGNN~\eqref{deep-GGNN} can be considered as a cascade of subsystems, so it is ISS if every subsystem is ISS. 
\end{proof}
\noindent The $\delta$ISS condition is more complex since there does not exist, to our knowledge, a general study in the literature for a cascade of $\delta$ISS systems. However, in this case we can state
\begin{theorem}
\label{deep-delta-stability}
The deep GGNN is $\delta$ISS stable if $\mathcal{A}^i_{\delta} \leq 1$ for every layer $i \in \{ 1, \dots, M\}$, where $\mathcal{A}_{\delta}^i$ are defined like in Theorem~\ref{dISS_stab}.
\end{theorem}
The proof is reported in Appendix~\ref{deep-proof}.
$\delta$ISS is a preferable property for deep GGNN as well as for single layer networks, as this property guarantees that any two state trajectories will converge under the same inputs.

\section{GGNN STABILITY UNDER COMMUNICATION DELAY}
\label{ggnn-stability-delay}
The previous scheme of GGNN can represent any dynamics on a graph. However, it does not take into account the communication steps and the possible delay in the application of the support matrix $S$. In each communication step, performed with sampling time $T$, we considered that the nodes communicate the data used in the graph filters and, at each $T$, we obtain  the output of a GGNN layer computed on the data $[x(t-K),x(t-(K-1)),\dots,x(t)]$. We express the delayed system in~\eqref{system} as: 
\begin{equation}
\small
    \begin{cases}
    \tilde{\bm{q}} = \sigma(\tilde{A}_{S^t}(\bm{x}(t-K)) + \tilde{B}_{S^t}(\bm{u}(t-K)) + \bm{\tilde{b}}) \\
    \hat{\bm{q}} = \sigma(\hat{A}_{S^t}(\bm{x}(t-K)) + \hat{B}_{S^t}(\bm{u}(t-K)) + \bm{\hat{b}})  \\
    \bm{x}(t+1) = \sigma_c(\hat{q}\circ A_{S^t}(\bm{x}(t-K)) + \tilde{q} \circ B_{S^t}(\bm{u}(t-K)) + \bm{b}) \\
    \bm{y} = Y_{S^t}(\bm{x}(t-K) + \bm{b_y})
    \end{cases}
    \label{eq:delayed_system}
\end{equation}
where the delay of $K$ communication steps can be represented by a delay between the input, old of $K$ steps, and the updating rule of the state. Moreover, we must consider graph filters with dynamic support matrices, i.e. matrices that change between each communication step. Thus, the graph filters considered so far become
\begin{equation*}
\small
H_{S^t}(\bm{x}(t)) = \bmatrix I_N \\ \vdots \\ \displaystyle \prod_{\tau=t-(K-1)}^{t} S(\tau) \endbmatrix^T diag(\bmatrix \bm{x}(t) \\ \vdots \\ \bm{x}(t-K) \endbmatrix) \bmatrix H_0 \\ \vdots \\ H_K \endbmatrix 
\end{equation*}
This filter expression is called unit-delayed filter~\cite{gama2022synthesizing}. Under Assumption \ref{assumption2}, the infinite norm of the support matrices in time is upper bounded. Therefore, the formulation~\eqref{eq:delayed_system} allows us to conclude that the system remains $\delta$ISS under the same conditions of the theorem~\ref{dISS_stab}.
\begin{remark}
As noted in~\cite{gama2022synthesizing}, the trajectories and the underlying graph observed at training time and the one observed at deployment are different, causing an increase of the training error. This is not an issue since, at a reasonable sampling time, the sequence of support matrices will not present drastic changes as their spectral characteristics are similar and, under Theorem~\ref{dISS_stab}, the graph filters satisfy~\eqref{eq:GNN-stability} with $C=(||H||\infty -1)||\bar{S}||_{\infty}/2$, and thus are stable to graph perturbations.
\end{remark}   
\section{VALIDATION EXAMPLES}
We experimentally confirmed the stability condition on two distributed control scenarios: flocking control and multi-robot motion control. Despite their simplicity, these scenarios have been previously studied ~\cite{tolstaya2020learning, I-GamaRibeiro2020GNNPathPlanning} using different approaches such as GNN and LSTM, allowing for a direct comparison with stable GGNN sGGNN). We also explicitly compare a sGGNN which satisfies the condition in Theorem~\ref{dISS_stab} and a GGNN that does not satisfy either Theorem~\ref{ISS_stab} or \ref{dISS_stab}. The stability condition is imposed by the following regularization in addition to the training loss:
\begin{equation}
    \Pi = \sum_{i=0}^{L} \rho_{-}\text{min}(0, \delta \mathcal{A}_{i} - 1 - \epsilon) + \rho_{+} \text{max}(0, \delta \mathcal{A}_{i} - 1 - \epsilon).
    \label{eq:reg_stab}
\end{equation}
This regularization applies the condition in the Theorem~\ref{dISS_stab} for each layer of GGNN with $ 0 < \rho_- \ll \rho_+$. By choosing $\rho_-$, $\Pi$ enforces fast convergence pushing $\delta \mathcal{A}_i$ toward zero at the cost of less accuracy, since the condition in Theorem~\ref{dISS_stab} is only a sufficient condition. For this reason, the choice of $\rho_-$ must be designed in the training process to achieve a desirable estimation error. In our implementation, we used $\rho^+ = 1$ and $\rho^-=0.01$.

\subsection{Flocking Control Example}
\label{flocking-control-example}
\begin{figure}
    \centering
    \includegraphics[scale=0.4]{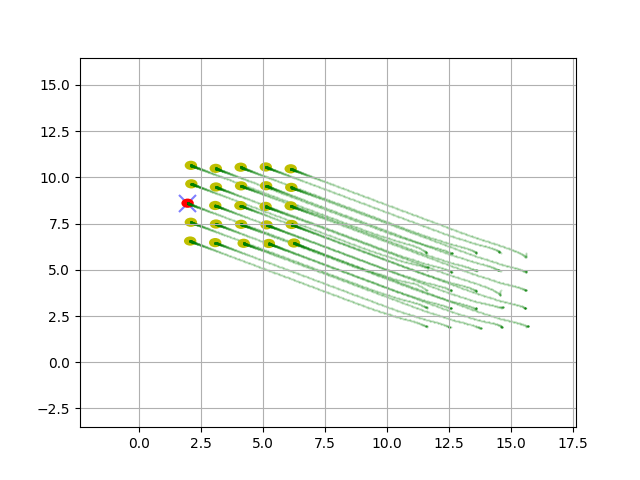}
    \caption{\textbf{Flocking control}: a group of agents (yellow dots) move in order to reach the same velocity and to avoid collision. The leader (red dot) moves in order to reach the target (blue cross) and avoids the collision with the other agents.}
    \label{fig:flocking_example}
\end{figure}
In the following, we show an application of the sGGNN on a case study involving flocking control (Fig.~\ref{fig:flocking_example}) with a leader. In the problem of flocking, the agents are initialized to follow random velocities while the goal is to have them all fly at the same velocity while avoiding collisions with each other. Moreover, one of the agents takes the leader role, conducting the team toward a target unknown to the other agents. Flocking is a canonical problem in decentralized robotics~\cite{yu2010distributed, beaver2021overview}.
\subsubsection*{\textbf{Dynamics and Expert Controller}}
We considered $N$ agents described by the position $\bm{r}(t) \in \mathbb{R}^{N \times 2}$ and the velocity $\bm{v}(t) \in \mathbb{R}^{N \times 2}$ with a double integrator dynamics
\begin{equation*}
        \bm{r}(t+1) = \bm{r}(t) + T\bm{v}(t); \quad 
        \bm{v}(t+1) = \bm{v}(t) + T\bm{u}(t);
\end{equation*}
with the discrete acceleration $\bm{u}(t) \in \mathbb{R}^{N \times 2}$ taken as system input.  Note that the agent dynamics is used for building the dataset and for simulation purpose, but it is not provided to the learning algorithm. \\
Since the objective is to make all the agents reach the same velocity, the control must be tuned in order to minimize the following cost function
\begin{equation}
    J(v(t)) = \frac{1}{N} \sum_{i=1}^N ||v_i(t) - \frac{1}{N}\sum_{j=1}^N v_j(t) ||^2_2
    \label{flocking-objective}
\end{equation}
where $v_i, v_j$ are the velocities for the agent $i$ and $j$, respectively. The cost function measures the distance of the agent velocities from the average velocities of the team. The cost $J(v(t))$ under the control $\bm{u}(t)$ can be analysed on the time horizon $T$ such that $t\in[0,T]$ to evaluate the convergence rate of the system. Moreover, the leader, randomly picked among the agents, does not follow the objective in~\eqref{flocking-objective} but rather it minimizes its distance from the target ($\bm{d}$). Therefore, the expert controller~\cite{tolstaya2020learning} for the followers is given by 
\begin{subequations}
\begin{equation}
    \bm{u}_f(t) = -\bm{L}(t)\bm{v}(t) - \nabla_{\bm{r}}CA(\bm{r}(t),\bm{r}_j(t))|_{j=1\dots N} 
\end{equation}
\text{and for leader it is}
\begin{equation}
    \bm{u}_l(t) = -W_p(\bm{r}_l(t) - \bm{d}(t)) - \nabla_{\bm{r}_l}CA(\bm{r}_l(t),\bm{r}_j(t))|_{j=1\dots N} 
\end{equation}
\label{eq:expert_flocking}
\end{subequations}
\noindent where $W_p$ is a gain, $\bm{r}_l \in \mathbb{R}^{2}$ is the leader position, $\nabla_{\bm{r}} CA(\bm{r}(t), \bm{r}_j(t))$/$\nabla_{\bm{r}_l}CA(\bm{r}_l(t), \bm{r}_j(t))$ are the gradient of the collision avoidance potential with respect to the position of the agents/leader $\bm{r}$/$\bm{r}_l$, evaluated at the position $\bm{r}(t)$/$\bm{r}_l(t)$ and the position of every other agent $\bm{r}_j(t)$ at time $t$. The i-element of $\nabla_{\bm{r}}CA$ for each robot $i$ with respect to robot $j$ is given by \cite{tanner2003stable}
\begin{equation}
    \nabla_{\bm{r}_i} CA(\bm{r}_{ij}) = \begin{cases}
    -\frac{\bm{r}_{ij}}{||\bm{r}_{ij}||_2^4} - \frac{\bm{r}_{ij}}{||\bm{r}_{ij}||_2^2} & if ||\bm{r}_{ij}||_2^2 \leq R_{CA} \\
    \qquad \bm{0} & otherwise
    \end{cases}
\end{equation}
with $\bm{r}_{ij} = \bm{r}_i - \bm{r}_j$ and $R_{CA} > 0$ indicating the minimum acceptable distance between agents. This potential function is a non negative, non smooth function that goes to infinity when the distance reduces and grows when the distance exceeds $R_{CA}$, in order to avoid the team losing the connectivity~\cite{tanner2003stable}. $\bm{u}_f(t),\bm{u}_l(t)$ are a centralized controller since computing them requires agent $i$ to have instantaneous evaluation of $\bm{L}(t)\bm{v}(t)$ and $\bm{r}_j(t)$ of every other agent $j$ in the team. $R_{CA}$ and $W_p$ are tunable parameters of the controllers.  

\subsubsection*{\textbf{Neural Network Architecture}}
We assume that the agents form a communication graph when they are in a sphere of radius $R$ between each others and that exchanges occur at the sampling time $T$, so that the action clock and the communication clock coincide. \\
The input features vector $\bm{w}_i \in \mathbb{R}^{10}$ of the robot $i$ for the designed neural network is 
\begin{equation}
\begin{split}
    \bm{w}_i = \Bigg[ & \bm{v}_i, \sum_{j \in \mathcal{NS}_i} \frac{\bm{r}_{ij}}{||\bm{r}_{ij}||_2^4}, \sum_{j \in \mathcal{NS}_i}\frac{\bm{r}_{ij}}{||\bm{r}_{ij}||_2^2}, \\ & \{\bm{0}_2, \bm{r}_l - \bm{d} \}, \{ [0,1],[1,0] \} \Bigg]
\end{split}
\end{equation}
where $\mathcal{NS}_i$ is the set of the sensing agents within a sphere of radius $R_{CA}$ centred in the robot $i$. Moreover, the vector contains the zero vector $\bm{0}_2 \in \mathbb{R}^{1 \times 2}$ and the one-hot encoding $[0,1]$, if the agent is a follower while $(\bm{r}_l - \bm{d})$ and $[1,0]$, if the agent is a leader. We chose the one-hot encoding instead of the binary one, because it allows differentiating the neural network weights between the leader and the follower. Note that all the information in the vector $\bm{w}_i$ are locally available at the sampling/control time $T$. \\
The core of the neural network for the flocking control is a layer of GGNN with $F=50$ features in the hidden state and filters length $K=2$. Note that the choice of $K$ affects the complexity of the stability condition imposed since it will constraint more parameters. The input features are first processed by a cascade of two fully connected layers of $128$ nodes before feeding the graph neural network. A readout of two layers with $128$ nodes combines the $F$-features GGNN hidden state to get the bidimensional control $\bm{u}$ saturated to the maximum admissible control. The input layers and the readout that encapsulate the GGNN shape a more realistic setting to test the stability of the graph neural network that is usually used in combination with other kinds of neural models. Following the Remark \ref{remark-nL}, we used normalized Laplacian as a support matrix 

\subsubsection*{\textbf{Training}}
\begin{figure*}[t]
\begin{subfigure}[t]{0.33\textwidth}
    \centering
    \includegraphics[scale=0.4]{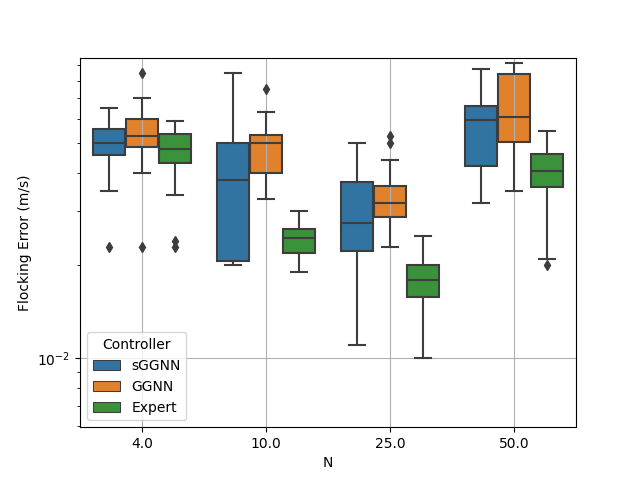}
    \caption{\centering Flocking error: variable team dimensions }
    \label{fig:flocking_error_N}
\end{subfigure}
\begin{subfigure}[t]{0.33\textwidth}
    \centering
    \includegraphics[scale=0.4]{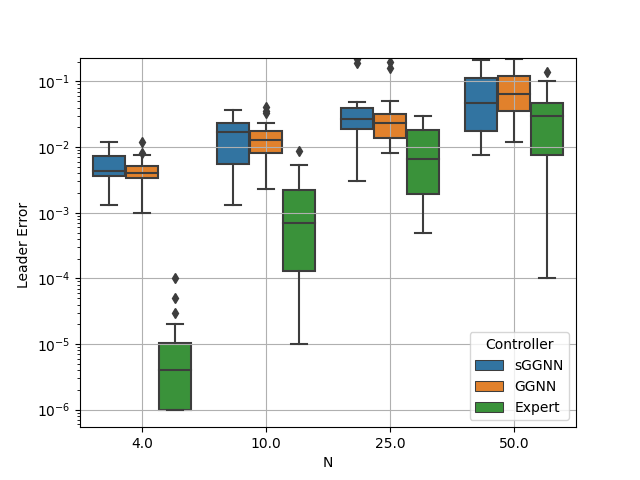}
    \caption{\centering Leader error: variable team dimensions}
    \label{fig:leader_error_N}
\end{subfigure}
\begin{subfigure}[t]{0.33\textwidth}
         \centering
         \includegraphics[scale=0.4]{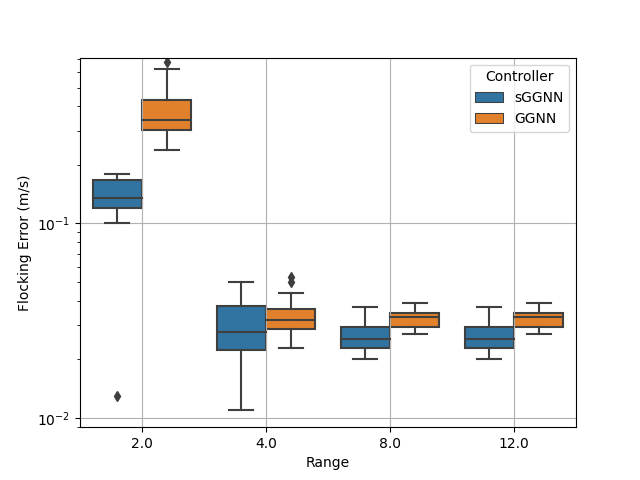}
    \caption{\centering Flocking error:  variable communication range}
    \label{fig:flocking_error_range}
    \end{subfigure}
     \hfill
     \begin{subfigure}[t]{0.33\textwidth}
         \centering
         \includegraphics[scale=0.4]{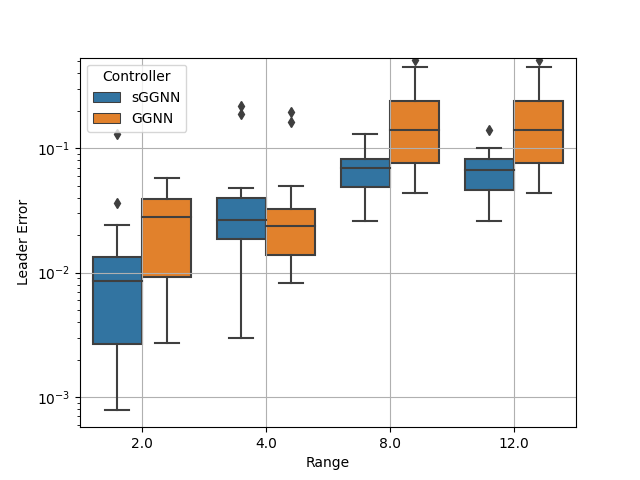}
        \caption{\centering Leader error: variable communication range}
    \label{fig:leader_error_range}
     \end{subfigure}
     \begin{subfigure}[t]{0.33\textwidth}
         \centering
         \includegraphics[scale=0.4]{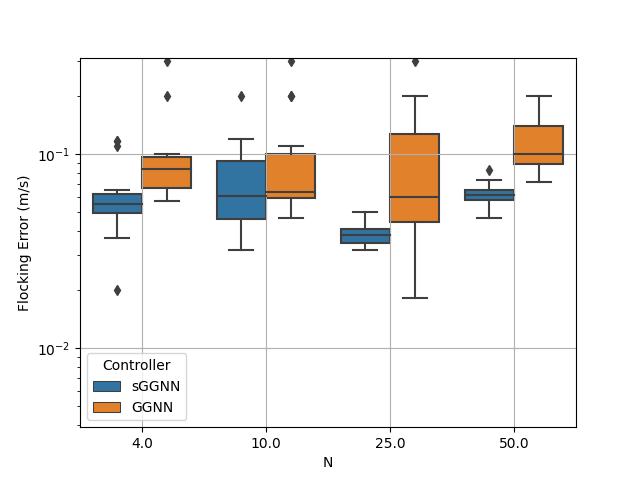}
    \caption{\centering Flocking error: network with communication delay and variable team dimensions}
    \label{fig:flocking_error_communication}
     \end{subfigure}
     \begin{subfigure}[t]{0.33\textwidth}
         \centering
         \includegraphics[scale=0.4]{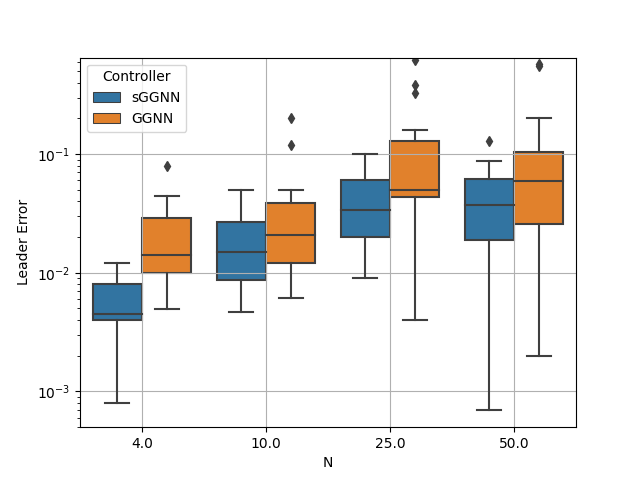}
    \caption{\centering Leader error: network with communication delay and variable team dimensions}
    \label{fig:leader_error_communication}
     \end{subfigure}
\caption{Flocking and Leader Error for stable (sGGNN) and non-stable GGNN controllers, varying the team size N with fixed communication range of $4$m (with and without instantaneous communication) and the communication range with $N=25$, reported using box plots that display median, minimum, maximum, $25$th/$75$th percentiles, and outliers.}
\label{fig:flocking_results}
\end{figure*}
\begin{figure}
\begin{subfigure}[t]{0.5\textwidth}
         \centering
         \includegraphics[scale=0.4]{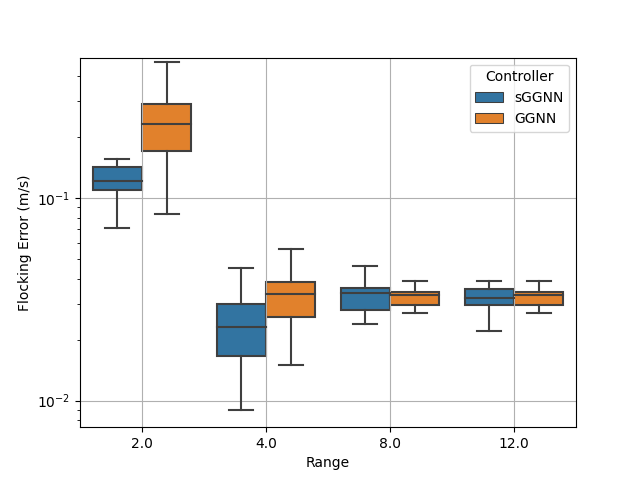}
    \caption{Flocking error}
    \label{fig:flocking_laplacian_error_communication}
     \end{subfigure}
     \begin{subfigure}[t]{0.5\textwidth}
         \centering
         \includegraphics[scale=0.4]{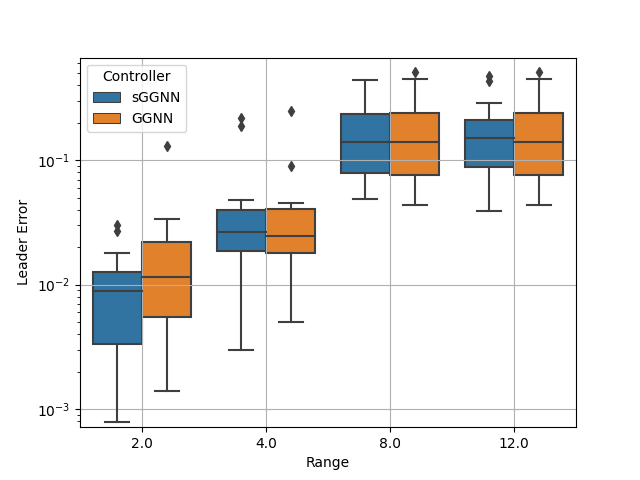}
    \caption{Leader error }
    \label{fig:leader_laplacian_error_communication}
     \end{subfigure}
\caption{ Flocking Error and Leader Error for stable and non-stable GGNN controllers with variable communication range using Laplacian}
\label{fig:flocking_laplacian_results}
\end{figure}
We collected a dataset by recording 120 trajectories,  further separated into three subsets of training, validation and test set using the proportion $70\%-10\%-20\%$, respectively. Each trajectory is generated by randomly positioning the agents in a square such that their inter-distance is between $0.6$~m and $1.0$~m and their initial velocities are picked at random from the interval $[ -2, 2 ]$~m/s in each direction. The leader is randomly selected among the agents and the target position is randomly located within a square of length $20$~m centered at the location of the leader. Regardless of the target location, the trajectories have a duration of $2.5$~s and input saturation at $5$~m/s$^2$. Moreover, the $120$ trajectories are recorded with a random number of agents among $N=[4,6,10,12,15]$. We fixed the communication range to $R=4$~m and the sensing to $R_{CA}=1$~m. We trained the models for $120$ epochs and executed the DAGGER algorithm~\cite{ross2011reduction} every $20$ epochs. The algorithm evaluates the expert controller in~\eqref{eq:expert_flocking} on the enrolled state trajectories applying the learned control and adding them to the training set. Note that, thanks to the use of DAGGER, we do not need a large dataset. We solve the imitation learning problem using the ADAM algorithm~\cite{kingma2014adam} with a learning rate $1e-3$ and forgetting factors $0.9$ and $0.999$. The loss function used for imitation learning is the mean squared error between the output of the model and the optimal control action.
\subsubsection*{\textbf{Results}}
In Fig.~\ref{fig:flocking_results}, we show a comparison between the stable GGNN (sGGNN) and non-stable GGNN controller for the flocking controller case. We evaluate the two controllers on 3 sets of experiments with $40$ trajectories each. In the experiments, we varied team size, communication range and network delay to test the robustness of the controllers. Figures~\ref{fig:leader_error_N},~\ref{fig:leader_error_range} and~\ref{fig:leader_error_communication} report the leader position error evaluated after a fixed time of $2.5$~s with respect to the leader starting location, i.e. $e_f/e_s$ with $e_f,e_s$ respectively being the final and the initial square distance of the leader from the target. Figures~\ref{fig:flocking_error_N},~\ref{fig:flocking_error_range} and~\ref{fig:flocking_error_communication} show the average flocking error~\eqref{flocking-objective} in the interval $0-2.5$~s in logarithmic scale. Moreover, we consider a failure when the control leads to an agent-agent collision, the leader-target distance diverges, or any agent-agent distance diverges (i.e. the team splits). 
In the first experiment, controllers' scalability is evaluated for team sizes $N=[4,10,25,50]$ while maintaining a fixed communication range of $4$ m and no communication delay. Figs.~\ref{fig:flocking_error_N} and~\ref{fig:leader_error_N} show sGGNN's improvement in achieving the flocking state, as seen by the average error, which is generally closer to the expert controller than GGNN. As the number of agents increases, inter-agent collision avoidance leads to low flocking error due to motion constraints for distributed and centralized controllers. However, in the case of $50$ agents, the leader struggles to drive the team toward the target due to the group cohesion, resulting in higher flocking and leader errors for all controllers. In general, sGGNN and GGNN show similar leader errors, even if, on average, this error is $7\%$-$12\%$ smaller for the non-stable neural network, at the cost of a higher flocking error.

In the second set of experiments, sGGNN demonstrates enhanced robustness when communication range differs from the training range $R=4$~m, as can be seen in Figs.~\ref{fig:flocking_error_range},~\ref{fig:leader_error_range}, where the range varies between $[2,4,8,10]$ and the team size is set at $N=25$ under instantaneous communication. Specifically, when the range is $R=2$~m, GGNN causes a flocking error of $0.40$~m/s with outliers up to $0.68$~m/s, greater than the average of $0.13$~m/s of the sGGNN. Moreover, with GGNN, we experienced more group division and a consequent splitting of the team, which explains the increase in the flocking error. However, in this condition, we also report the leader divergence and a consequent worse leader error with respect to the stable control of $27\%$. While the stable controller never results in a failure situation, the non-stable GGNN causes $12\%$ failures with a range of $R=2$~m. As expected, when the communication range increases, the flocking errors decrease for both controllers, since they are able to communicate with more agents at the same time. However, in this case, the leader error increases, since the leader is often encapsulated by the other team agents and it is thus forced to follow them to not collide, causing a slower convergence to the target. Note that this behavior also affects the expert controller. 

In the last set of experiments, we consider not instantaneous communication (delay of $T=0.01$~s) and evaluate again the transference at scale. With large team sizes, the non-stable controller fails $20\%$-$35\%$ of the experimented trajectories, while the stable one succeeded $100\%$. Figures~\ref{fig:flocking_error_communication},~\ref{fig:leader_error_communication} show the consistency of the results with respect to the non-delay case, confirming that the controller remains stable even with not instantaneous communication. In some cases, sGGNN shows a high error variance.  This behavior can be seen, for example, in Fig~\ref{fig:flocking_error_N} with N=$10$, and in Fig~\ref{fig:flocking_error_range} with R=$4$m, where the sGGNN errors boxes are large around the mean value. However, sGGNN error is overall lower and remains consistently  closer to the expert controller one.

In Fig.~\ref{fig:flocking_laplacian_results}, we also reported the performances of the sGGNN and GGNN trained and tested with the Laplacian matrix to confirm what we stated in the Remark \ref{remark-nL}. As we can see, with a higher communication range than the one used at training time, the differences between the two controllers disappear except for differences caused by training errors. Even if the condition in Theorem 2 is only sufficient and possibly conservative, from the results, it is clear that imposing stability over the distributed controller improved the agent performances compared to scenarios where stability was not imposed.
\subsection{Multi Robot Motion Control Example}
\label{multi-robot-motion-planning-example}
\begin{figure}[t]
    \centering
    \includegraphics[scale=0.55]{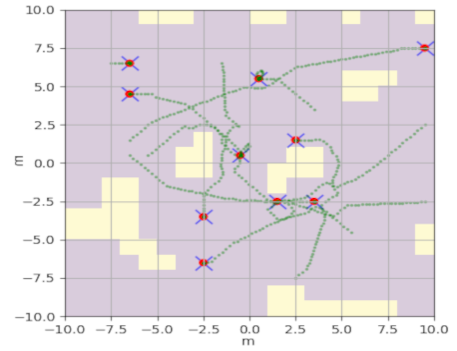}
    \caption{\textbf{Multi Robot Motion Control}: a group of agents (red dots) move to reach their targets (blue cross) avoiding agent-agent and agent-obstacle (in yellow) collisions.}
    \label{fig:motion_planning}
\end{figure}
In this section, we address multi-robot motion in a cluttered space (Fig.~\ref{fig:motion_planning}), a problem already studied in~\cite{gupta2017cooperative,le2019multi,tordesillas2021mader}. Our approach is based on~\cite{li2020graph}, which aims at solving the multi-robot path planning using GNN. In contrast to this work, we consider smooth trajectories and continuous space instead of discrete space and decisions. This choice highlights the robustness and stability properties of GGNN when controlling robots with smooth inputs. The objective is to guide a group of robots, initially located at random positions, towards their individual targets in the cluttered space.
~\subsubsection*{\textbf{Dynamics and Expert Controller}}
We consider $N$ agents described by the position $\bm{r}(t) \in \mathbb{R}^{N \times 2}$ in a single integrator dynamics with the velocity $\bm{u}(t) \in \mathbb{R}^{N \times 2}$ taken as the system input. As before, the learning algorithm is agnostic to the agent dynamics.
For the expert controller, we used a combination of RRT \cite{gammell2014informed} to find an obstacle-free path and MPC to control the agents in a continuous time. The MPC minimizes the divergence of the agents from the RRT-generated paths while constraining the agent motion to avoid inter-agent and obstacle collisions: 
\begin{equation*}
\begin{aligned}
& \underset{\bm{u}(t),\bm{r}(t)}{\text{min}}
& & \sum_{i=0}^N||\bm{r}_{id}(t)||^2_2 \\
& \quad \text{s.t.} & & \dot{\bm{r}}_i = \bm{u}_i \; i = 1, \ldots, N. \\
& & & g_o(\bm{r_{io}}(t)) \geq \bm{b}_o, \; i = 1, \ldots, N. \\
& & & g_{ij}(\bm{r_{ij}}(t)) \geq \bm{b}_{ij}, \; i,j = 1, \ldots, N. \\
& & & || \bm{u}_i ||_2 \leq 1 \; i = 1, \ldots, N.
\end{aligned}
\end{equation*}
where $\bm{r_{id}}$ is the distance from the RRT-path for the agent $i$, $g_o, g_{ij}$ are the quadratic distances between the agent positions and the obstacles in the space ($\bm{r_{io}}$) and between the agents ($\bm{r_{ij}}$). As we can see, the MPC is naturally a centralized solution, since it uses all the agent dynamics to generate the control inputs.  

\subsubsection*{\textbf{Neural Network Architecture}}
As in the previous example, we assume the agents to form a communication graph if they are within a communication radius of $R$. The input features vector $\bm{w}_i \in \mathbb{R}^{10}$ of the robot $i$ for the designed neural network is 
\begin{equation}
\begin{split}
    \bm{w}_i = \Bigg[ & \bm{r}_{id}, \sum_{o \in \mathcal{NS}_{io}} \frac{\bm{r}_{io}}{||\bm{r}_{io}||_2^4}, \sum_{o \in \mathcal{NS}_{io}}\frac{\bm{r}_{io}}{||\bm{r}_{io}||_2^2}, \\ & \sum_{j \in \mathcal{NS}_i} \frac{\bm{r}_{ij}}{||\bm{r}_{ij}||_2^4}, \sum_{j \in \mathcal{NS}_i}\frac{\bm{r}_{ij}}{||\bm{r}_{ij}||_2^2}, \Bigg]
\end{split}
\label{eq:input-multim}
\end{equation}
where $\mathcal{NS}_i, \mathcal{NS}_{io}$ are respectively the set of the agents and obstacles in the sensing range $R_C$ of the robot $i$. At time $T$, all this information is available to the agent $i$.
Similarly to the previous case, we use a 2-layer MLP with 128 nodes for input and 2 layers with 128 nodes for readout to control the agent's velocity. However, for testing deep-GGNN stability, we employ a 2-layer GGNN with 30 features in the hidden state and a filter of length K=2. The support matrix used here is the normalized Laplacian.

\subsubsection*{\textbf{Training}}
We recorded 40 trajectories to build the dataset, further separated in three subsets of training, validation and test set using the proportion $70\%-10\%-20\%$, respectively. The trajectories are the results of RRT+MPC controller running with $N=10$ agents randomly located in a square space of $20$m $\times 20$m with $15\%$ obstacles density. The agent targets are equally randomly located in the free space. We fixed the communication range to $R=4$m and the sensing one to $R_C=1$m. The training runs for $200$ epochs with the DAGGER algorithm executed every $20$ epochs. We used the ADAM algorithm with learning rate $\num{1e-3}$ and forgetting factors $0.9$ and $0.999$. The loss function used for imitation learning is the mean squared error between the output of the model and the expert control action.
\begin{figure*}[t]
\begin{subfigure}[t]{0.33\textwidth}
    \centering
    \includegraphics[scale=0.4]{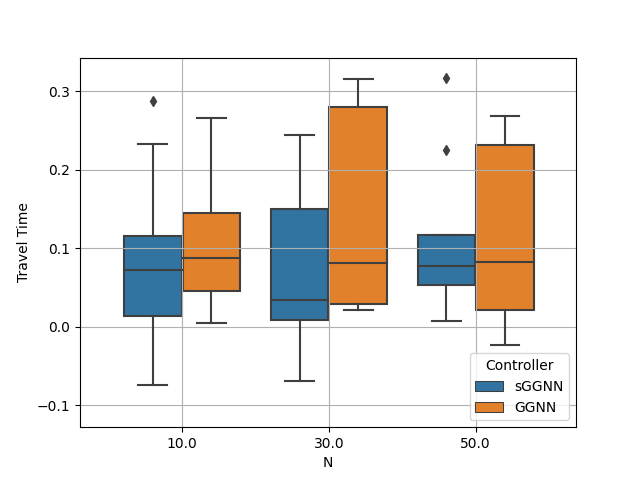}
    \caption{Travel time - variable team dimensions}
    \label{fig:multim_travel_time_N}
\end{subfigure}
\begin{subfigure}[t]{0.33\textwidth}
    \centering
    \includegraphics[scale=0.4]{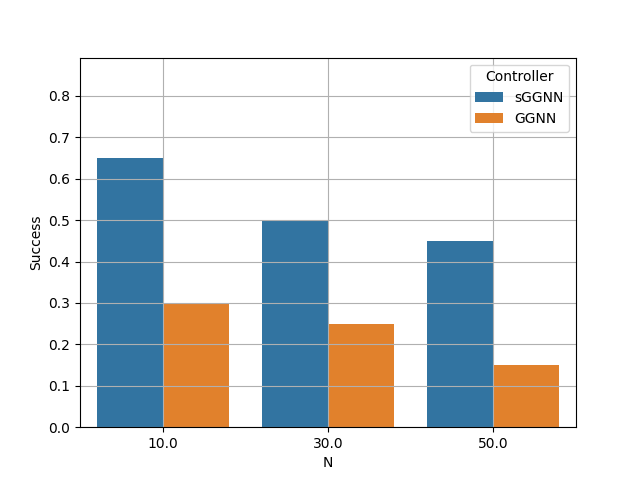}
    \caption{Success rate - variable team dimensions}
    \label{fig:multim_success_N}
\end{subfigure}
\begin{subfigure}[t]{0.33\textwidth}
         \centering
         \includegraphics[scale=0.4]{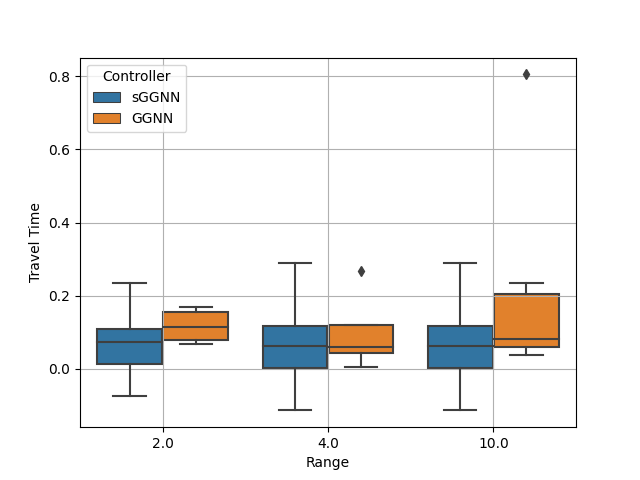}
    \caption{Travel time - variable communication range}
    \label{fig:multim_travel_time_range}
    \end{subfigure}
     \hfill
     \begin{subfigure}[t]{0.33\textwidth}
        \centering
        \includegraphics[scale=0.4]{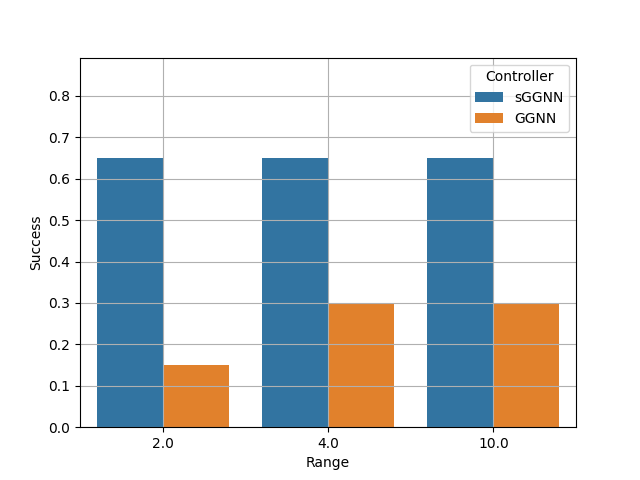}
        \caption{Success rate - variable communication range}
        \label{fig:multim_success_range}
    \end{subfigure}
    \begin{subfigure}[t]{0.33\textwidth}
        \centering
        \includegraphics[scale=0.4]{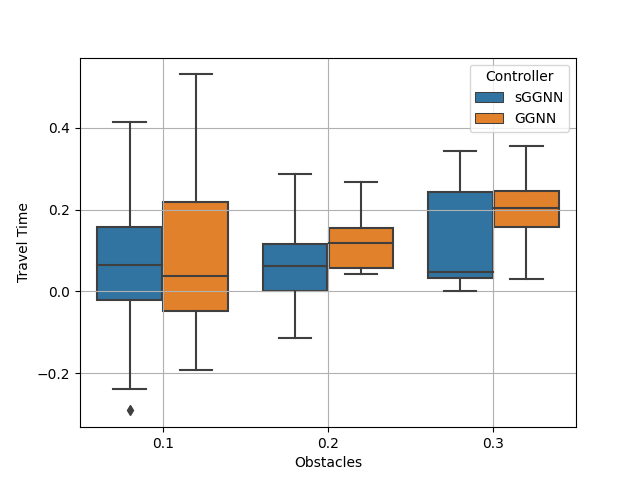}
        \caption{Travel Time - variable obstacle density}
        \label{fig:multim_travel_time_obstacles}
     \end{subfigure}
     \begin{subfigure}[t]{0.33\textwidth}
         \centering
         \includegraphics[scale=0.4]{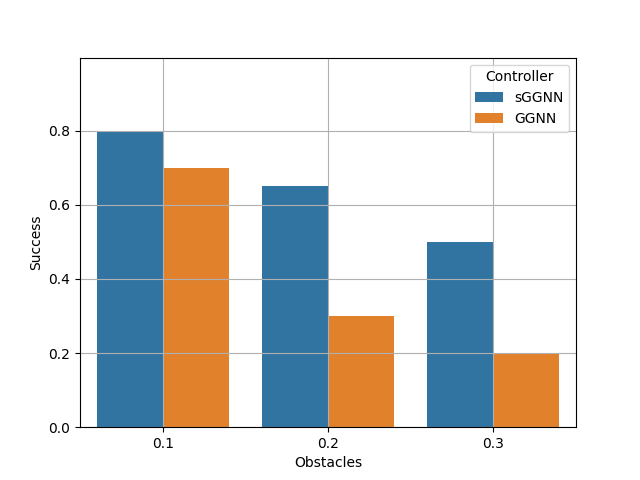}
        \caption{Success rate - variable obstacle density}
        \label{fig:multim_success_obstacles}
     \end{subfigure}
\caption{ Success rate and flow time for stable and non-stable GGNN controllers evaluation varying the team size, the communication range and the obstacles density for a $20m \times 20m$ map; the flow time increasing is computed as $(Tf-Tf^*)/Tf^*$ with $Tf^*$ expert controller time of arrival.}
\label{fig:multimotion_results}
\end{figure*}
\begin{figure*}[t]
    \begin{subfigure}[t]{0.5\textwidth}
        \centering
        \includegraphics[width=7cm,height=5cm]{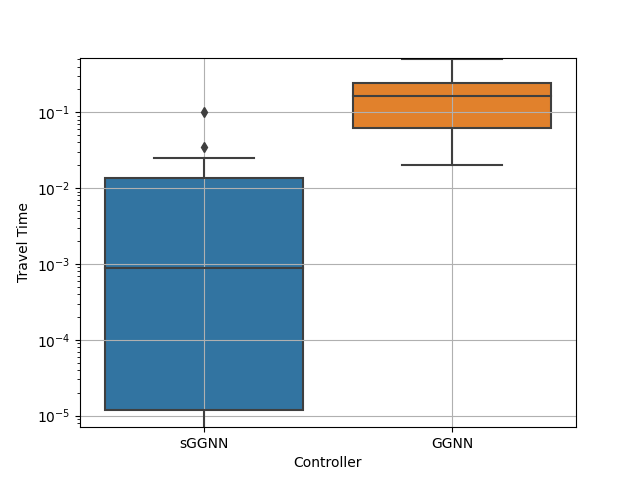}
        \caption{Flow Time - GGNN + rrt}
        \label{fig:multim_travel_time_rrt}
    \end{subfigure}
    \begin{subfigure}[t]{0.5\textwidth}
         \centering
         \includegraphics[width=7cm,height=5cm]{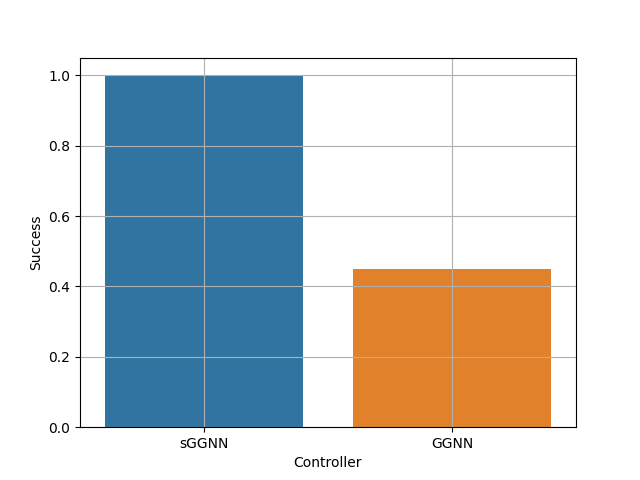}
        \caption{Success rate - GGNN + rrt}
        \label{fig:multim_obstacles_rrt}
     \end{subfigure}
\caption{Success rate and flow time for stable and non-stable GGNN controllers following a precomputed rrt path to avoid the obstacles. The controllers run on the map of $20m \times 20m$ with $30\%$ of randomly generated obstacles and 50 agents. The flow time increasing is computed as $(Tf-Tf^*)/Tf^*$ with $Tf^*$ expert controller time of arrival.}
\label{fig:multimotion_rrt}
\end{figure*}
\subsubsection*{\textbf{Results}} We evaluate the stability condition for the multi-robot motion control and report the results in Fig.~\ref{fig:multimotion_results}. The comparison is carried out on three sets of experiments evaluating transferable at scale, robustness on communication range and obstacle density. We recorded $40$ trajectories for each case on a $20$m $\times 20$m map. For this application, we are primarily interested in the control success rate showed in Figs.~\ref{fig:multim_success_N},~\ref{fig:multim_success_range} and~\ref{fig:multim_success_obstacles}, where we consider successful the trajectories free of collisions and deadlocks. On successful trajectories, we also computed the travel time increase with respect to the expert controller. We obtained all the trajectories in the non-instantaneous communication setting and, when not stated otherwise, with team size, the communication range and the obstacle density respectively of $N=10, R=4.0$m and $20\%$. 

As in the flocking example, we first test the scalability of the two controllers to the team size ($N$) variations among $[10,30,50]$. sGGNN has a success rate that attests between $65\%$ and $45\%$, showing better performances compared to the GGNN one, that is always below $30\%$ (see Fig.~\ref{fig:multim_success_N}). The travel time increase is comparable for the two control solutions as we can see in the Figs.~\ref{fig:multim_travel_time_N},~\ref{fig:multim_travel_time_range} and ~\ref{fig:multim_travel_time_obstacles}, even if the stable one presents more situations of negative travel time increasing. This latter may have values below zeros due to the presence of the RRT in the expert controller that finds an obstacle-free but not the shortest path. Hence GGNNs control can lead to trajectories that are faster since they work with the target location directly. This phenomenon is less evident when the number of robots (Fig.~\ref{fig:multim_travel_time_N}) or the obstacle density increases (Fig.~\ref{fig:multim_travel_time_obstacles}).
The sGGNN does not show particular robustness to variations in the obstacle density since, as reported in Fig.~\ref{fig:multim_success_obstacles}, the successful rate of $80\%$ for $10\%$ of obstacles drops to $45\%$ for $30\%$ of obstacles, the same success rate of the $N=50$ and $20\%$ of obstacles, even if it generalizes better than the non-stable learned controller as confirmed by the higher success rate and the lower average flow time with respect to the non-stable controller. This suggests a poor obstacle avoidance behaviour, further demonstrated by the results in Fig.~\ref{fig:multim_obstacles_rrt} where we isolate failures caused by agent-agent collision by testing a combination of gated GNN and RRT. For an obstacle density of $30\%$ and 50 agents, we feed the RRT path to neural models projecting it on the sensing range by replacing the target location in the input layer \eqref{eq:input-multim}. In this case, sGGNN+RRT reaches $100\%$ success and an average travel time of $10$e$-3$, confirming that sGGNN realizes a well-distributed approximation of the MPC. On the contrary, GGNN+RRT still results in a high failure rate due to agent-agent collisions.
Varying the communication range, as shown in Fig.~\ref{fig:multim_success_range}, produces comparable results, wherein sGGNN upholds a steady success rate. In contrast, GGNN encounters heightened instances of collisions among agents when communication ranges are reduced. This outcome underscores the resilience of sGGNN to fluctuations in graph connectivity levels. When communication ranges are lower, robots tend to establish disjointed graphs more easily. Additionally, the performance remains unchanged since agents in close proximity form connected sub-graphs, allowing the neural network's internal state to quickly adapt and control the agent effectively under the new conditions.
\section{CONCLUSIONS}
\label{conclusions}
In this work, we devise sufficient conditions for the ISS and Incremental ISS of gated graph neural networks. When GGNN are used to learn distributed policies, the proposed stability conditions allow to guarantee that the trained networks enjoy the ISS/$\delta$ISS property, which is particularly useful during the synthesis of distributed controllers. The proposed condition has been tested on the flocking control and multi robot motion control, showing good modelling performances. Results suggest that enforcing stability properties on the learned controller makes it closer to the expert centralized one and more robust to parametric changes in a deployment scenario, such as communication radius and team size.
\appendices
\section{}
\subsection{proof to Theorem~\ref{ISS_stab}}
\label{proof_ISS}
In the following, we report the proof of  theorem~\eqref{ISS_stab}.
\proof
Without loss of generality, we assume that the initial state belongs to the invariant set $\mathcal{X} = [-1,1]^{N\times F}$. This is not a restrictive assumption since even if $\bm{x}(0) \notin \mathcal{X}$, at the next iteration it will be in $\mathcal{X}$ due to the activation function $\sigma_c$. In light of the definitions given in~\eqref{eq:definitions}, it holds that  
\begin{flalign}
    \begin{aligned}
    ||\bm{x}^+||_{\infty} & \leq || \sigma_c (\hat{q} \circ A_S(\bm{x}) + \tilde{q} \circ B_S(\bm{u}) + \bm{b})||_{\infty} \\ & \leq || \hat{q} \circ A_S(\bm{x}) + \tilde{q} \circ B_S(\bm{u}) + \bm{b}||_{\infty} \\ & \leq  \sigma_{\hat{q}} ||S_K||_\infty || A ||_{\infty} ||\bm{x} ||_{\infty} + \\ & \qquad \sigma_{\tilde{q}} ||S_K||_\infty ||B ||_{\infty} ||\bm{u}||_{\infty} + ||\bm{b}||_\infty\\ & \leq \mathcal{A} ||\bm{x}||_{\infty} + \mathcal{B} || \bm{u}||_{\infty} + ||\bm{b}||_\infty 
    \end{aligned}
    \label{state_ineq}
\end{flalign}
From theorem~\eqref{ISS_stab}, by iterating the~\eqref{state_ineq} for $t$ steps we get 
\begin{equation}
    \begin{aligned}
    ||\bm{x}(t)||_{\infty} & \leq \mathcal{A}^t ||\bm{x}(0)||_{\infty} + \sum_{k=1}^{t} \mathcal{A}^t(\mathcal{B}||\bm{u}||_{\infty} + ||\bm{b}||_\infty) \\ 
    ||\bm{x}(t)||_{\infty} & \leq \mathcal{A}^t ||\bm{x}(0)||_{\infty} + (1-\mathcal{A})^{-1}\mathcal{B} ||\bm{u}||_{\infty} + \\ & \qquad (1-\mathcal{A})^{-1}||\bm{b}||_\infty 
    \end{aligned}
\end{equation}
which proves the ISS property according to the definition~\ref{ISS_def}

\endproof
\subsection{proof to Theorem~\ref{dISS_stab} }
\label{prrof_dISS}
In the following, we report the proof of theorem~\ref{dISS_stab}
\proof
Given two states $\bm{x}_1, \bm{x}_2$, it holds that  
\begin{equation*}
    \begin{aligned}
    \bm{x_1}^+ - \bm{x_2}^+= & \sigma_c(\hat{q}_1 \circ A_{S1}(\bm{x_1}) + \tilde{q}_1 \circ B_{S1}(\bm{u_1}) + \bm{b})  \\ & - \sigma_c(\hat{q}_2 \circ A_{S2}(\bm{x_2}) + \tilde{q}_2 \circ B_{S2}(\bm{u_2}) + \bm{b}).
    \end{aligned}
\end{equation*}
Owing to the Lipschitz assumption of $\sigma_c$ and $\sigma$ for Lipschitz constants respectively of $1$ and $\frac{1}{4}$, the distance between the two state trajectories is bounded by  
\begin{equation}
    \begin{aligned}
    ||\bm{x_1}^+ & - \bm{x_2}^+||_{\infty} \leq \\ & || (\hat{q}_1 \circ A_{S1}(\bm{x}_1) + \tilde{q}_1 \circ B_{S1}(\bm{u}_1)) - \\ & \quad (\hat{q}_2 \circ A_{S2}(\bm{x}_2) + \tilde{q}_2 \circ B_{S2}(\bm{u}_2)) ||_{\infty} \leq \\ & || (\hat{q}_1 \circ A_{S1}(\bm{x}_1)  - \hat{q}_2 \circ A_{S2}(\bm{x}_2)) + \\ & \quad (\tilde{q}_1 \circ B_{S1}(\bm{u}_1) - \tilde{q}_2 \circ B_{S2}(\bm{u}_2)) ||_{\infty} \leq \\ & ||\hat{q}_1 \circ (A_{S1}(\bm{x_1}) - A_{S2}(\bm{x_2}))||_{\infty} + \\ & \quad ||(\hat{q}_1 - \hat{q}_2) \circ A_{S2}(\bm{x_2})||_{\infty} + \\ &  \quad ||\tilde{q}_1 \circ (B_{S1}(\bm{u_1}) - B_{S2}(\bm{u_2}))||_{\infty} + \\ & \quad ||(\tilde{q}_1 - \tilde{q}_2) \circ B_{S2}(\bm{u_2})||_{\infty}.
    \end{aligned}
    \label{eq:state_diff}
\end{equation}
Different input features will correspond to different graph topologies and different support matrices $S_1,S_2$; thus different graph filters with the same parameters. For this reason, the differences of graph filters in the previous equation require further development. Let us focus on the state dependent graphs of the previous inequality. It holds that
\begin{equation}
    \begin{aligned}
    ||\hat{q}_1 \circ & (A_{S1}(\bm{x_1}) - A_{S2}(\bm{x_2}))||_{\infty} \leq \\ & \sigma_{\hat{q}} ||S_{K1} ( I_{K} \otimes \bm{x_1})A - S_{K2} ( I_{K} \otimes \bm{x_2})A ||_{\infty} \leq \\ & \sigma_{\hat{q}} || S_{K1} ( I_{K} \otimes \bm{x_1} - I_{K} \otimes \bm{x_2}) + \\ & \quad (S_{K1} - S_{K2})(I_{K} \otimes \bm{x_2})||_{\infty}||A||_{\infty}    .
    \end{aligned}
    \label{eq:state_ineq}
\end{equation}
Under the assumption of theorem~\ref{dISS_stab}, we have $||S_{K1}||_{\infty},||S_{K2}||_{\infty} \leq ||\bar{S}_{K}||_{\infty}$. In light of $||\bm{x}||_{\infty} \leq 1$, equation~\eqref{eq:state_ineq} becomes
\begin{equation*}
    \begin{aligned}
    ||\hat{q}_1 \circ (A_{S1}(\bm{x_1}) - & A_{S2}(\bm{x_2}))||_{\infty} \leq \\ & \sigma_{\hat{q}} ( ||\bar{S}_{K}||_{\infty}||A||_{\infty}||\bm{x_1} - \bm{x_2}||_{\infty} + \\ & (||S_{K1}-S_{K2}||_{\infty})||A||_{\infty}).
    \end{aligned}
    \label{eq:state_ineq_final}
\end{equation*}
Applying the same reasoning to the other terms in the inequality~\eqref{eq:state_diff}, we obtain
\begin{equation*}
        \begin{aligned}
        ||\bm{x_1}^+ &- \bm{x_2}^+||_{\infty} \leq \\ & 
        (\sigma_{\hat{q}} ||\bar{S}||_\infty ||A||_{\infty}+\frac{1}{4}||\bar{S}||^2_\infty||\hat{A}||_{\infty} ||A||_{\infty} + \\ & \frac{1}{4} ||\bar{S}||^2_\infty||\tilde{A}||_{\infty} ||B||_{\infty})||\bm{x_1} - \bm{x_2}||_{\infty} + \\ & ( \sigma_{\tilde{q}}||\bar{S}||_\infty||B||_{\infty} + \frac{1}{4}||\bar{S}||^2_\infty||\hat{B}||_{\infty} ||A||_{\infty} + \\ & \frac{1}{4}||\bar{S}||^2_\infty||\tilde{B}||_{\infty}||B||_{\infty})||\bm{u_1} - \bm{u_2}||_{\infty} + \\ & \mathcal{W}(||S_{K1}-S_{K2}||_{\infty}) \leq \\ &
        \mathcal{A}_{\delta} ||\bm{x_1} - \bm{x_2}||_{\infty} + \mathcal{B}_{\delta} ||\bm{u_1} - \bm{u_2}||_{\infty} + \\ & \mathcal{W}(||S_{K1}-S_{K2}||_{\infty})
        \end{aligned} 
\end{equation*}
where $\mathcal{W}$ gathers all the coefficient multiplying the difference $||S_{K1}-S_{K2}||_{\infty}$. We can consider this latter as an additional bounded input, $||S_{K1} - S_{K2}||_{\infty} \leq ||\bar{S}_{K}||_{\infty} - 1$ which, analogously to the input features, is defined by the team state. Hence, as stated in the theorem~\eqref{dISS_stab}, it holds 
\begin{equation}
    \begin{aligned}
    ||\bm{x_1}(t) &- \bm{x_2}(t)||_{\infty} \leq \mathcal{A}^t_{\delta} ||\bm{x_1}(0) - \bm{x_2}(0)||_{\infty} + \\ & (1-\mathcal{A}_{\delta})^{-1} \delta \mathcal{B} ||\bm{u_1} - \bm{u_2}||_{\infty} + \\ & (1-\mathcal{A}_{\delta})^{-1} \mathcal{W}||S_{K1}-S_{K2}||_{\infty}.
    \end{aligned}
    \label{eq:incremental_trajectories}
\end{equation}
The state trajectories have then a maximum distance that is asymptotically bounded by a function monotonically increasing with the maximum distance between the input sequences for
\begin{equation}
    \gamma_{\delta} = \bmatrix (1-\mathcal{A}_{\delta})^{-1}\delta \mathcal{B} & (1-\mathcal{A}_{\delta})^{-1}\mathcal{W} \endbmatrix \Vmatrix \bm{u_1} - \bm{u_2}  \\ S_{K1}- S_{K2} \endVmatrix_{\infty}.
    \label{eq:gamma_dISS}
\end{equation} 
Therefore the system is incrementally ISS under the definition~\ref{dISS_def}. 

\endproof

\subsection{proof of Theorem~\ref{deep-delta-stability}}
\label{deep-proof}

\begin{proof}
To analyse the incremental stability it is useful to separate each layer. From the proof of theorem ~\ref{prrof_dISS}, for the first layer we know that:
\begin{equation}
\begin{aligned}
    ||\bm{x_1}^{1+} & - \bm{x_2}^{1+}||_{\infty} \leq \mathcal{A}^1_{\delta} ||\bm{x^1_1} - \bm{x^1_2}||_{\infty} + \mathcal{B}^1_{\delta} ||\bm{u_1} - \bm{u_2}||_{\infty} + \\ & \mathcal{W}^1||S_{K1}-S_{K2}||_{\infty} 
\end{aligned}
\end{equation}
As a result, for the second it holds 
\begin{equation}
    \begin{aligned}
        ||\bm{x_1}^{2+} & - \bm{x_2}^{2+}||_{\infty} \leq \\ & \mathcal{A}^2_{\delta} ||\bm{x^2_1} - \bm{x^2_2}||_{\infty} + \mathcal{B}^2_{\delta} ||\bm{x_1}^{1+} - \bm{x_2}^{1+}||_{\infty} + \\ & \mathcal{W}^2||S_{K1}-S_{K2}||_{\infty} \leq \\ & \mathcal{A}^2_{\delta} ||\bm{x^2_1} - \bm{x^2_2}||_{\infty} + \mathcal{B}^2_{\delta} \mathcal{A}^1_{\delta} ||\bm{x_1}^{1} - \bm{x_2}^{1}||_{\infty} + \\ & \mathcal{B}^2_{\delta} \mathcal{B}^1_{\delta} ||\bm{u_1} - \bm{u_2}||_{\infty} + \\ & \mathcal{B}^2_{\delta} \mathcal{W}^1||S_{K1}-S_{K2}||_{\infty} + \mathcal{W}^2(||S_{K1}-S_{K2}||_{\infty}  
    \end{aligned}
\end{equation}
Denoting $\Delta X = \bmatrix \bm{x^1_1} - \bm{x^1_2} \quad \dots \quad \bm{x^M_1} - \bm{x^M_2} \endbmatrix^T$, $\Delta U = \bm{u_1} - \bm{u_2}$, $\Delta S_K = S_{K1} - S_{K2} $ iterating the same reasoning for $M$ layers we get 
\begin{equation}
    \begin{aligned}
    ||\Delta X^+||_{\infty} \leq & \begin{bmatrix}
    A^1_{\delta} & 0 & \dots & 0 \\
    B^2_{\delta} A^1_{\delta} & A^2_{\delta} & \dots & 0 \\
    \vdots & \ddots & \ddots & \vdots \\
    A^1_{\delta} \prod\limits_{h=2}^{M} B^h_{\delta} & \dots & \dots & A^M_{\delta}
    \end{bmatrix} || \Delta X ||_{\infty} + \\  \begin{bmatrix}
    B^1_{\delta} \\
    B^2_{\delta}  \\
    \vdots \\
    \prod\limits_{h=1}^{M} B^h_{\delta}
    \end{bmatrix} ||\Delta & U||_{\infty} + \begin{bmatrix}
    \mathcal{W}^1 \\
    \mathcal{W}^1 B^2_{\delta} + \mathcal{W}^2  \\
    \vdots \\
    \prod\limits_{h=1}^{M-1} \mathcal{W}^h B^{h+1}_{\delta} + \mathcal{W}^M  
    \end{bmatrix} ||\Delta S||_{\infty} \\ \leq M_{\delta}|| \Delta X ||_{\infty} +& M_{B\delta}||\bm{u_1} - \bm{u_2}||_{\infty} +  M_{\mathcal{W}\delta}||S_{K1}-S_{K2}||_{\infty}
    \end{aligned}
\end{equation}
Since the matrix $M_{\delta}$ is lower triangular, its eigenvalues are the elements of its diagonal $\mathcal{A}^i_{\delta}$ and are equal to the eigenvalues of the matrix $M_{\delta}$.
\end{proof}
\bibliographystyle{IEEEtran} 
\bibliography{IEEEabrv,bibliografy}
\begin{IEEEbiography}[{\includegraphics[scale=0.3, width=1in,height=1in,clip,keepaspectratio]{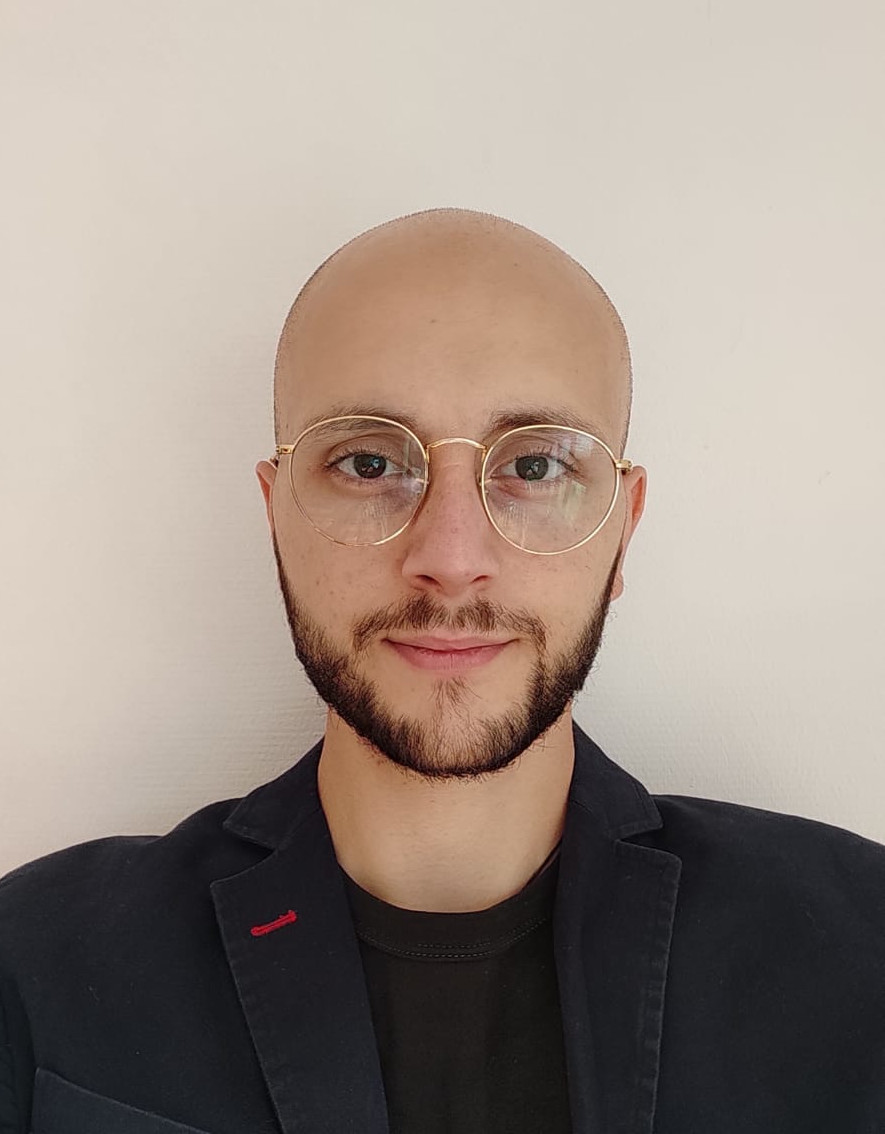}}]{Antonio Marino} obtained his M.Sc.~degree in robotics engineering at the University of Genova in 2020. A. Marino is currently a PhD candidate in the Rainbow team at IRISA/Inria Rennes working on learning-based multi-robot formation control.
\end{IEEEbiography}
\begin{IEEEbiography}[{\includegraphics[scale=0.4, width=1in,height=1in,clip,keepaspectratio]{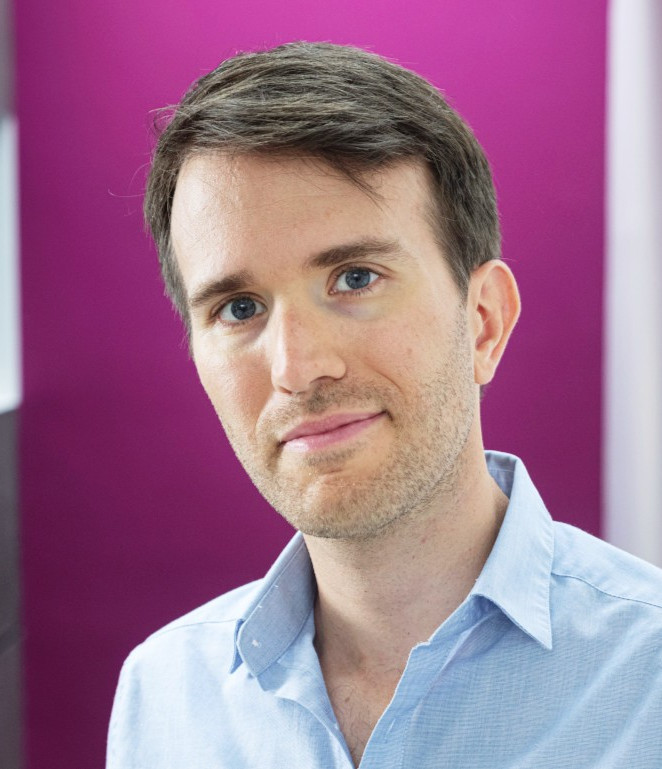}}]{Claudio Pacchierotti} (SM'20) is a
tenured researcher at CNRS-IRISA in Rennes, France, since 2016. He was previously a postdoctoral researcher at the Italian Institute of Technology, Genova, Italy. Pacchierotti earned his PhD at the University of Siena in 2014. He was Visiting Researcher in the Penn Haptics Group at Univ. Pennsylvania in 2014, the Dept. of Innovation in Mechanics and
Management at Univ. Padua in 2013, the Institute for Biomedical Technology and Technical Medicine (MIRA) at University of Twente in 2014, and the Dept. Computer, Control and Management Engineering of the Sapienza Univ. Rome in 2022. Pacchierotti received the 2014 EuroHaptics Best PhD Thesis Award and the 2022 CNRS Bronze Medal. He is Senior Chair of the IEEE Technical Committee on Haptics, Co-Chair of the IEEE Technical Committee on Telerobotics, and Secretary of the Eurohaptics Society.
\end{IEEEbiography} 

\begin{IEEEbiography}[{\includegraphics[scale=0.3,width=1.2in,height=1.2in,clip,keepaspectratio]{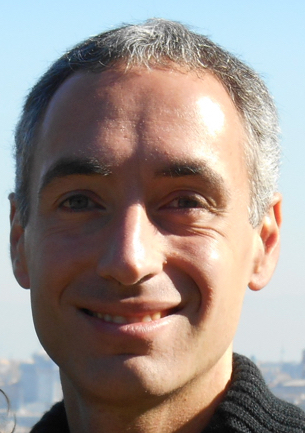}}]{Paolo Robuffo Giordano}  (M'08-SM'16) received his M.Sc. degree in Computer Science Engineering in 2001, and his Ph.D. degree in Systems Engineering in 2008, both from the University of Rome ``La Sapienza''. In 2007 and 2008 he spent one year as a PostDoc at the Institute of Robotics and Mechatronics of the German Aerospace Center (DLR), and from 2008 to 2012 he was Senior Research Scientist at the Max Planck Institute for Biological Cybernetics in T\"ubingen, Germany. He is currently a senior CNRS researcher head of the Rainbow group at Irisa and Inria in Rennes, France.
\end{IEEEbiography}

\end{document}